\documentclass[manuscript]{acmart}
\usepackage{longtable,booktabs} 
\usepackage[ruled]{algorithm2e} 
\usepackage{enumerate}
\usepackage{pifont}
\newcommand{\cmark}{\ding{51}}%

\begin{document}
\title{A Systematic Review of Automated Grammar Checking in English Language}

\author{Madhvi Soni}
\affiliation{%
  \institution{Jabalpur Engineering College}
  \streetaddress{Department of Computer Science \& Engineering}
  \city{Jabalpur}
  \state{M.P.}
  \postcode{482011}
  \country{India}}
\email{madhvi.soni21@gmail.com}

\author{Jitendra Singh Thakur}
\affiliation{%
  \institution{Jabalpur Engineering College}
  \streetaddress{Department of Computer Science \& Engineering}
  \city{Jabalpur}
  \state{M.P.}
  \postcode{482011}
  \country{India}}
\email{jsthakur@jecjabalpur.ac.in , jsthakur@iiitdmj.ac.in}

\renewcommand\shortauthors{Madhvi Soni et al}

\begin{abstract}
Grammar checking is the task of detection and correction of grammatical errors in the text. English is the dominating language in the field of science and technology. Therefore, the non-native English speakers must be able to use correct English grammar while reading, writing or speaking. This generates the need of automatic grammar checking tools. So far many approaches have been proposed and implemented. But less efforts have been made in surveying the literature in the past decade. The objective of this systematic review is to examine the existing literature, highlighting the current issues and suggesting the potential directions of future research. This systematic review is a result of analysis of 12 primary studies obtained after designing a search strategy for selecting papers found on the web. We also present a possible scheme for the classification of grammar errors. Among the main observations, we found that there is a lack of efficient and robust grammar checking tools for real time applications. We present several useful illustrations- most prominent are the schematic diagrams that we provide for each approach and a table that summarizes these approaches along different dimensions such as target error types, linguistic dataset used, strengths and limitations of the approach. This facilitates better understandability, comparison and evaluation of previous research.
\end{abstract}


\maketitle

\section{Introduction}

English is a West Germanic language which is the second most common language of the world. Over 600 million speakers use English as a second language (ESL) or English as a foreign language (EFL). While writing text in their second or foreign language, people might make errors. Therefore, it is essential to be able to detect these grammar errors and correct them as well. Grammar checking by a human becomes inconvenient at times such as when human resource is limited, the size of the document is large or the grammar checking is to be done on a regular basis. Therefore, it would be beneficial to automate the process of grammar checking. A grammar checking tool can provide automatic detection and correction of any faulty, unconventional or controversial usage of the underlying grammar.\\

The trend of developing such tools has been evolved from 80's till now. Earliest grammar checking tools (e.g., Writer's Workbench\cite{macdonald1983human}) were aimed at detecting punctuation errors and style errors. In 90's, many tools were made available in the form of commercialized software packages (e.g., RightWriter\cite{neuman1991rightwriter}). In recent decades, rapid development has been seen in this field. For example, Park et al \cite{park1997english} developed a grammar checker as a web application for university ESL students, Tschumi et al\cite{tschichold1997developing} developed a tool aimed at French native speakers writing in English, Naber developed an tool named LanguageTool \cite{naber2003rule} to detect a variety of English Grammar errors, Brockett et al \cite{brockett2006correcting} presented error correction using machine translation and Felice et al \cite{felice2014grammatical} presented a hybrid system. Existing approaches are hard to compare since most of their tools are not available. Moreover, they are developed on different datasets and targets detection of different types of errors. Study and comparative analysis of previous literature is important to gain future research directions, yet very few efforts have been put to survey grammar checking approaches in the last decade. Therefore, we are highly motivated to review the existing literature for identifying the related issues and concerns, and present them in a single study to our research community. \\

This paper reports on a systematic review \cite{keele2007guidelines} that focuses on various approaches for automatic detection and correction of grammar errors in English text. While reviewing the literature, we have tried to summarize as many details as possible, explaining the complete step by step workflow of the approach along with its strengths and limitations (if any).  Our intention is to provide a platform for comparing the existing approaches that will help in taking further research decisions. Also, we have searched the literature to find various types of errors, but found that all the researchers are addressing a set of errors that is different from each other. Thus, we identify major types of errors and suggest an error classification scheme based on a five point criteria. We explain these types of errors along with their demonstrative examples. To the best of our knowledge, our study is the first one of its kind.\\

The paper is organized into following sections: Section II presents the method of performing systematic review. This section describes our research questions, search strategy, paper selection criteria and method of data extraction from the selected papers. Section III presents our suggested scheme to classify various English grammar errors. Section IV presents the classification of grammar checking techniques. Section V presents a detailed review of various approaches whose results are significant in this field. Finally, section VI concludes our paper and suggests some directions for further research.

\section{Systematic Review Method}
A systematic literature review is a well-planned procedure to search, identify, extract from, analyze, evaluate and interpret the existing literature works that are relevant to a particular research interest \cite{yue2011systematic},\cite{keele2007guidelines}. A systematic review is different from a conventional review as it summarizes the existing work in a more complete and unbiased manner \cite{keele2007guidelines}. Systematic reviews are undertaken to sum up the existing approaches, identifying their limitations, suggesting further research directions, and to provide a background for new research actions \cite{keele2007guidelines}.\\

We report a systematic review on grammar checking in English language. As per the recommended guidelines \cite{keele2007guidelines}, we have adopted five necessary steps to carry this review. In the first step, we formulate the research questions that will be addressed by this systematic review.  In the second step, we design a strategy to search for the research papers online. Third step defines the paper selection criteria to identify relevant works. The fourth step is extraction of data from primary studies and finally, in the last step we examine the data.

\subsection {Research Questions:}

\begin {enumerate} [RQ1]
\item What are the different types of errors in English grammar?
\item How can we classify them? Is there a classification scheme in the literature?
\item What are the various techniques of grammar checking?
\item What are the strengths and limitations of these techniques?
\item	What are existing approaches of grammar checking? What are the methods they use?
\item	Is there any experiment conducted by the authors to evaluate the performance of the approach?
\item	If yes, what results have been obtained?
\item	What types of errors are detected and corrected by these approaches?
\item	How far these approaches are able to correctly identify the errors?
\item	Is there any tool support available?

\end{enumerate}

\subsection {Search Strategy:}
Our search strategy starts by defining a query string. To form the string, we identified three groups of search terms: population terms, intervention terms and outcome terms. 
\begin{itemize}
\item Population Terms:These are the keywords that represent the domain of research. (e.g., grammar checking, grammar correction, English grammar errors, types of errors, error classification, and ESL errors.)
\item Intervention terms: These are the keywords that represent the techniques applied on population to achieve an objective. (e.g., automatic detection, detect, detecting, automatic correction, correct, correcting and identification.)
\item Outcome terms: These are the related factors of importance. (e.g., better, faster, efficient and improved performance.)
\end{itemize}

We performed an exhaustive search on ``Google scholar" to identify the papers to be reviewed. Since the search resulted in collection of a large number of papers, it is necessary to identify only the useful papers that can answer our specific research questions. Thus, we applied inclusion/exclusion criteria to select papers that can serve as primary studies in this systematic review.

\subsection{Inclusion/exclusion criteria:}

Our inclusion/exclusion criteria are completely based on our previously defined research questions. For each paper, we read the paper's title and abstract to identify the relevant papers. Furthermore, full text was read to take the final decision. Following points were considered while deciding on the selection of primary studies:
\begin{itemize}
\item Papers irrelevant to the task of grammar checking are excluded.
\item Papers proposing grammar checking on languages other than English are completely ignored.
\item Papers describing types of errors made by native speakers of a specific language (e.g., errors made by only Arab writers) were excluded.
\item Papers that do not provide sufficient technical information of their approach were excluded. (e.g., \cite{mozgovoy2011dependency})
\item In case of approaches those participated in a shared task(CoNLL-2013 and 2014), we include only the best performing approach.
\end{itemize}
	
After the electronic search, a total of 113 papers were identified to investigate. 35 duplicates were eliminated and 36 papers were eliminated in the first round by reading the abstract and introduction. So, 42 papers were remaining for further investigation.  After reading full-text, 29 papers were eliminated and finally 1 more was eliminated \cite{mozgovoy2011dependency} due to lack of implementation details. Thus, we identified 12 primary studies.

\subsection{Data Extraction:}

For data extraction, we used a tabular format where each primary study is reviewed under table headings such as name of the approach, technique used, steps involved in the approach, types of the errors addressed by the approach, experiments conducted by the authors (if any), dataset used in the experiment, outcomes of the experiment, name of the software tool designed (if any), and strengths and shortcomings of the approach (if any). Later, content of this table is used to write a detailed review of each primary study.

\section{Types of Errors}
\begin{figure}[htbp]
	\centering
		\includegraphics[width=1.00\textwidth]{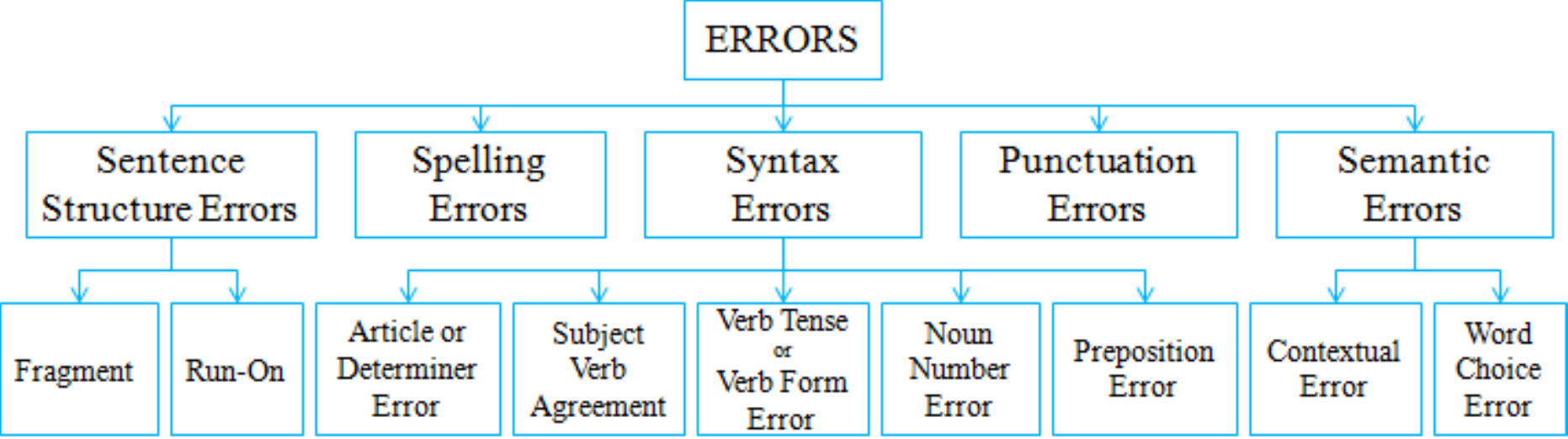}
	\caption{Our Suggested Scheme for Classification of Errors}
	\label{fig:errortypes}
\end{figure}

This section will address our research questions RQ1 and RQ2. Before actual implementation of any grammar checking approach, it is important to identify major types of errors and their classification on the basis of some criteria. For example, some researchers have classified the errors in the corpus based on whether they are automatically detectable or needs human assistance. Naber\cite{naber2003rule} classifies various errors into four types namely spelling errors, style errors, grammar (syntax) errors and semantic errors. Wagner et al\cite{wagner2007comparative} reports four types of errors namely agreement errors, real word spelling errors(contextual errors), missing word errors and extra word errors. Lee et al\cite{lee2008correcting} reports two types of errors namely syntax errors and semantic errors. Z Yuan in her doctoral thesis\cite{yuan2017grammatical} states five types of errors namely lexical errors, syntactic errors, semantic errors, discourse errors and pragmatic errors. Other than this, there is no general classification of grammar errors to the best of our knowledge. However an overview of major types of errors can be found in many web articles. Thus, we are highly motivated to suggest an error classification scheme. Please see figures~\ref{fig:comparison3} and ~\ref{fig:errortypes} for comparison of our scheme with previous schemes. \\

We have considered following points while designing our suggested classification scheme. 

\begin{itemize}

\item \textbf{Frequency of error:} More frequent errors should be kept in separate groups. For instance, five types of syntax errors are the most frequent errors that occur in ESL text\cite{rozovskaya2013university} so they are classified into separate groups. Similarly, spelling and punctuation errors are also very common. See figure~\ref{fig:ourscheme}(a).\\

\item \textbf{Validity of text:} Errors should be separated on the basis of how it makes the text invalid. For instance, syntax error invalidates a text due to violation of grammar rules. Similarly, sentence structure error invalidates a sentence due to violation of sentence structuring rules\cite{hornby1995guide} and a spelling error invalidates a word if it violates language orthography. See figure~\ref{fig:ourscheme}(b).\\

\item \textbf{Level of an error:} Some errors are detected at sentence level while others can be detected at word level i.e., taking two or three words. For instance, there is no need to check complete sentence to detect spelling errors. Similarly, checking words before and after a preposition would be sufficient to detect a preposition error, while fragments can be detected using parse tree pattern of a complete sentence. See figure~\ref{fig:ourscheme}(c).\\

\item \textbf {Nature of error:} The errors that are more annoying and difficult to detect should be separated from simpler ones. For instance, spelling error is rather formal which can easily be detected using a spell checker, while detection of a semantic error requires real-world knowledge. \\

\item \textbf{Error type overlap:} The error types in the classification scheme are overlapping. It cannot be completely avoided but we have tried to minimize it. For example, a run-on sentence can also be a punctuation error and a missing preposition error can also be a sentence structure error.\\
\end{itemize}

Again considering the frequency, nature and validity, we kept punctuation rules into a separate class of errors. Trying to minimize the overlapping, we reached to the final classification shown in figure~\ref{fig:errortypes}.

{{\centering
\begin{figure} [htpb]
\centering\includegraphics[width=\textwidth]{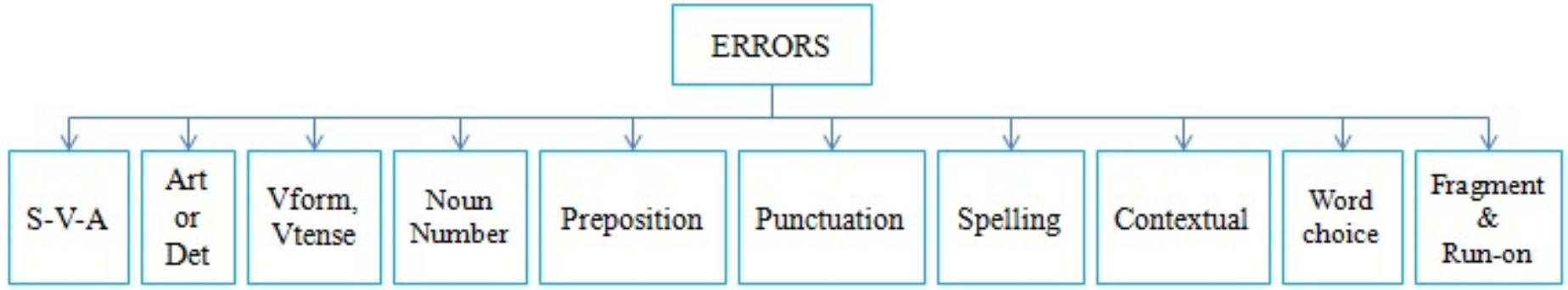} (a)
 \vspace{5pt} \linebreak 
\centering\includegraphics[width=\textwidth]{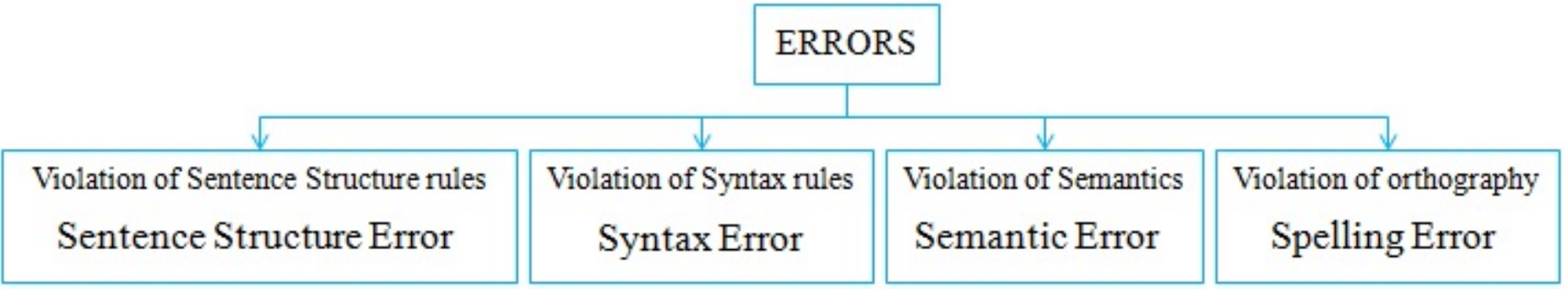} (b)
  \vspace{5pt} \linebreak 
\centering\includegraphics[width=\textwidth]{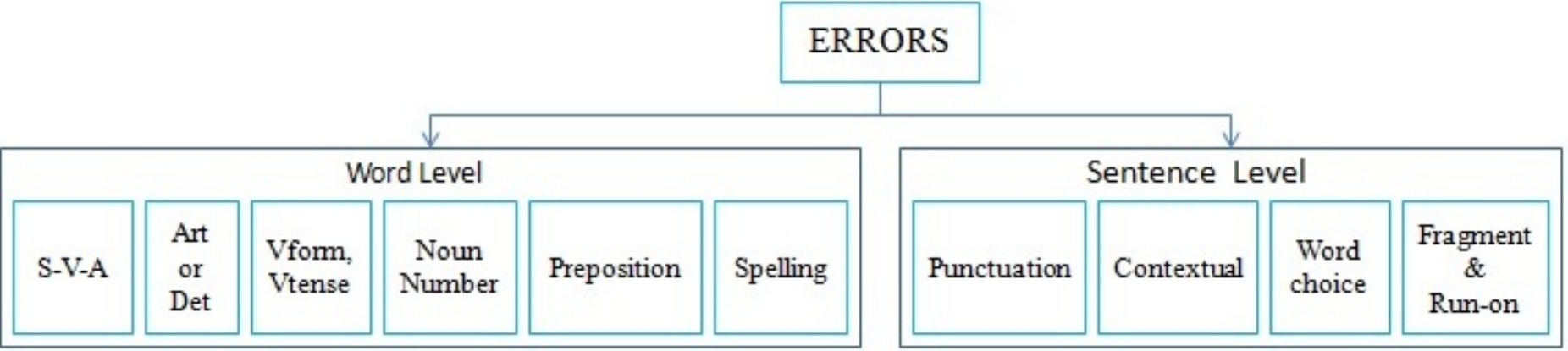} (c)
  \vspace{5pt} \linebreak 
 \centering\includegraphics[width=\textwidth]{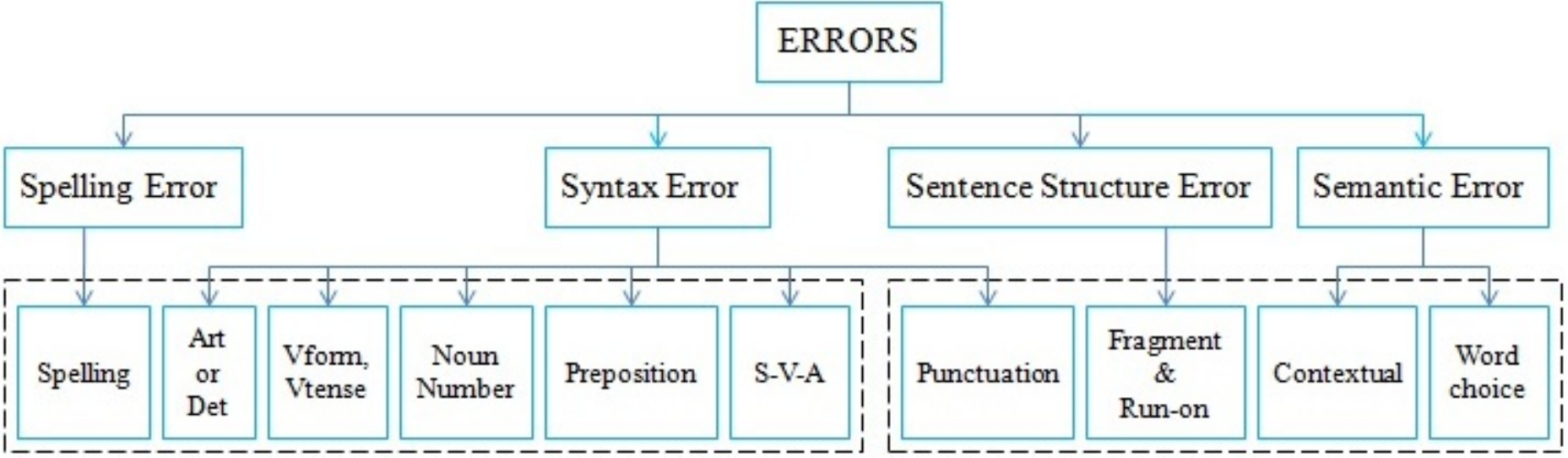}  (d) 
\caption{Classification of errors based on (a) frequency, (b) validity, (c) level and (d) combining (a), (b) and (c).}
\label{fig:ourscheme}
\end{figure}‎
}}

{\linespread{0.5} {\centering
\begin{figure}[htpb]‎
\includegraphics [height=4cm,width=0.87\textwidth]{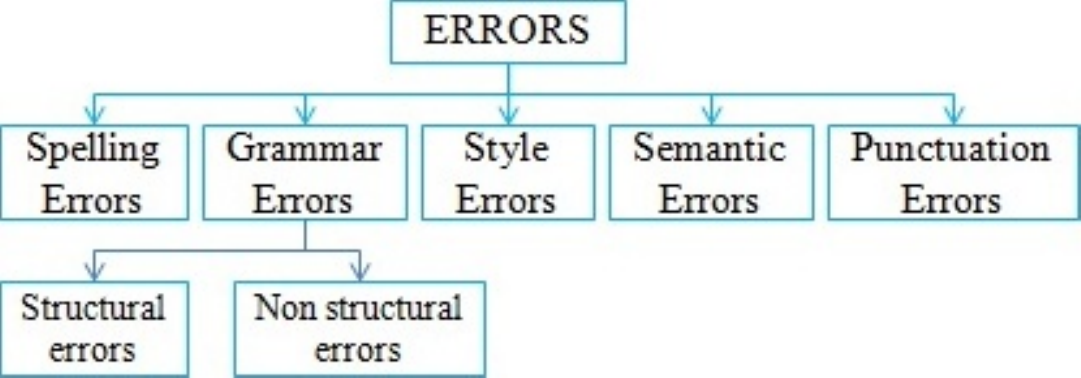}
\newline (a) \vspace{10pt} \linebreak 
\includegraphics [height=3.5cm, width=0.7\textwidth]{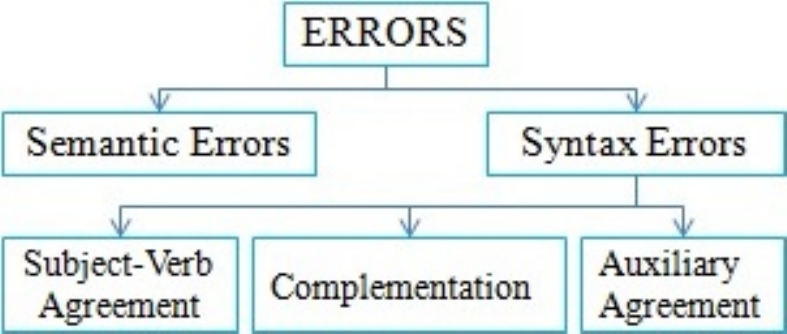}
\newline (b) \vspace{10pt} \linebreak 
\includegraphics [height=2.5cm, width=0.9\textwidth]{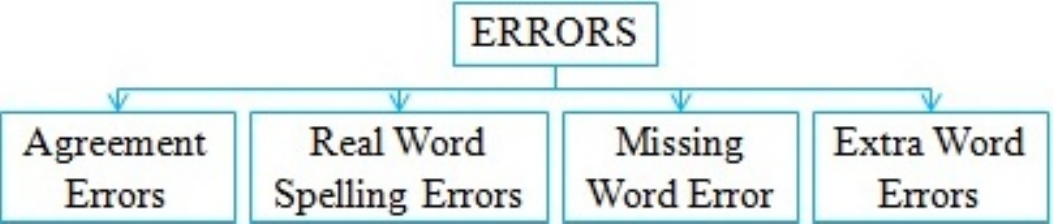}
\newline (c) \vspace{10pt} \linebreak 
\includegraphics [height=4cm, width=0.9\textwidth]{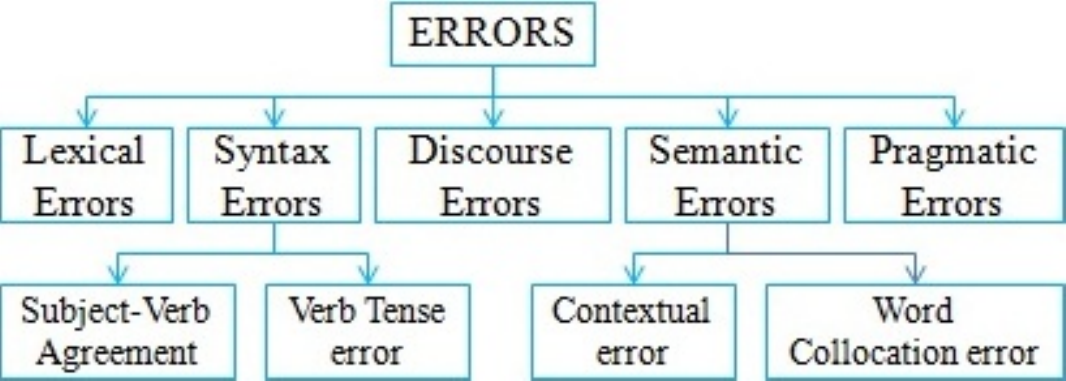} 
 \newline (d) \vspace{10pt} \linebreak 
\caption{Classification schemes given by (a) Naber\cite{naber2003rule}, (b) Lee et al\cite{lee2008correcting}, (c) Wagner et al\cite{wagner2007comparative}, (d) Z yuan\cite{yuan2017grammatical}}
\label{fig:comparison3}
\end{figure}‎
}}

Here, we are describing our suggested classification of errors. We give erroneous sentences for each type of error and their corrections are given in the bracket. All the examples have been taken from \cite{wren2000english}.

\begin{enumerate}
\item \textbf{Sentence Structure Error:} Sentence structure refers to the organization of different POS components within a sentence to give it a meaning. Structuring has a high impact on sentence's readability. Hornby\cite{hornby1995guide} has formulated 25 patterns of English sentences. If none of those patterns are found, the sentence can be considered as ill-formed or say erroneous. Such an ill-formed sentence can further be classified as \textbf{fragments} and \textbf{run-ons.} A fragment is an incomplete sentence in which either subject or verb is missing or it may be a sentence having dependent clause without the main clause \cite{yeung2015automatic}. A run-on sentence is two independent clauses missing a punctuation or necessary conjunction between them, which affects the readability of text. Sentence structure errors may contain other type of errors within them. Examples 1, 2 are correctly constructed while examples 3,4,5,6,7 are erroneous. Examples 4,5,6 are fragments while example 7 is a run-on.

\begin{flushleft} 
\textit{Example 1-} She began singing. (S-V-Gerund) \linebreak
\textit{Example 2-} She wants to go. (S-V-to-infinitive) \linebreak
\textit{Example 3-} She began to singing. (Misplaced `to' or `-ing') \linebreak
\textit{Example 4-} Wants to go. (Subject is missing) \linebreak
\textit{Example 5-} A fair little girl under a tree. (Verb is missing) \linebreak
\textit{Example 6-} Because he is ill. (`because' makes it a dependent clause, main clause is missing) \linebreak
\textit{Example 7-} I ran fast missed the train. (Conjunction `but' is missing) \linebreak 
\end{flushleft}

\item	\textbf{Punctuation Error:} Punctuation marks like comma, semi-colon, full stop etc. are used to separate sentence elements. A missing punctuation or unnecessary punctuation can alter the meaning of the sentence. Hence, it is important to detect and correct the punctuation errors in English text.

\begin{flushleft}
\emph{Example 8-} He lost lands money reputation and friends. (lands, money, reputation and friends) \linebreak
\emph{Example 9-} Alas she is dead ! (Alas ! She is dead.) \linebreak
\emph{Example 10-} How are you? Mohan? (How are you, Mohan?) \linebreak
\emph{Example 11-} Exactly so, said Alice. (``Exactly so,") \linebreak
\end{flushleft}

\item \textbf{Spelling Error:} Spelling error is the generation of a meaningless string of characters. A common reason for such errors is the typing mistakes done by the writers. These are the most common error types that can be found easily by any spell or grammar checking tool. Generally these tools have a list of known words. Any word outside this list is considered as a spelling error.

\begin{flushleft}
\emph{Example 12-} Death lays his icey hand on kings. (icy) \linebreak
\emph{Example 13-} Many are called, but few are choosen. (chosen) \linebreak 
\end{flushleft}

\item	\textbf{Syntax Error:} Any error violating the English grammar rules is called as syntax error. Syntax errors can be of many types depending upon the inherent relationship between the words of a sentence. Most grammar checkers aims at finding and detecting various types of syntax errors. Syntax errors can be subdivided into five subtypes: \newline

\begin{enumerate}
\item	\textbf{Subject-Verb Agreement Error:} A sentence written in English must have an agreement between subject and verb in terms of person and number. This agreement is shown in example 14 and 15.

\begin{flushleft}
\emph{Example 14-} He is not to blame. (subject-`he' ( third person singular)) (verb-`is' (third person singular)) \linebreak
\emph{Example 15-} They are not on good terms. (subject-`they' (third person plural)) (verb-`are'  (third person plural)) \linebreak 
\end{flushleft}

\item	\textbf{Article or Determiner Error:} This type of error occurs either when an article or determiner is missing in the sentence or when a wrong article or determiner is used.

\begin{flushleft}
\emph{Example 16-} Book you want is out of print. (The book) \linebreak
\emph{Example 17-} He returned after a hour. (an hour) \linebreak
\end{flushleft}

\item \textbf{Noun Number Error:}  In English, uncountable or mass nouns do not have plurals. So noun number error occurs when a plural form of uncountable nouns is used in the text.

\begin{flushleft}
\emph{Example 18-} He paid a sum of money for the informations. (information) \linebreak
\emph{Example 19-} The sceneries here are very good. (The scenery here is very good.) \linebreak
\end{flushleft}

\item \textbf{Verb Tense or Verb Form Error:} Verb tense or verb form conveys the time and state of the idea or event. This type of error occurs when a writer uses a different tense or form of verb from the intended one.

\begin{flushleft}
\emph{Example 20-} It is raining since yesterday. (has been raining) (`since' gives the idea that the event has started in the past and is still continuing) \linebreak
\emph{Example 21-} She leaves school last year. (left) (`last year' indicates a finished event of the past) \linebreak
\emph{Example 22-} The boys are play hockey. (playing) (the event is currently happening, so -ing form of verb is required) \linebreak
\end{flushleft}

\item \textbf{Preposition Error:} Prepositions are the words preceding a noun or pronoun, used to express a relation to other element in the clause. In literature, preposition errors are addressed separately because of the fact that it is difficult to master them.

\begin{flushleft}
\emph{Example 23-} He sat a stool. (He sat on a stool.) \linebreak
\emph{Example 24-} He has recovered of his illness. (from his illness) \linebreak
\end{flushleft}

\end{enumerate}

\item \textbf{Semantic Error:} The errors that do not violate English grammar rules, but make the sentence senseless or absurd, are called as semantic errors. A semantic error can be a \textbf{contextual error}\cite{bigert2004probabilistic} or \textbf{wrong word choice} error. When a wrongly typed word is a real word in the language, it is not detected as a spelling error, yet it does not fit in the given context; such errors are called as contextual errors.  Wrong word choice error is using a rare word (possibly due to limited knowledge of vocabulary) which is often not used in the given context. Examples 25,26 are contextual errors while 27,28 are word choice errors.

\begin{flushleft}
\emph{Example 25-} Our team is better then theirs. (`then' is not a spelling mistake, but the context gives an idea of comparison, indicating correct word as `than') \linebreak
\emph{Example 26-} The jury were divided in there opinions. (their opinions) \linebreak
\emph{Example 27-} A group of cattle is passing. (A herd of cattle) \linebreak
\emph{Example 28-} I am going to the library to buy a book. (use `bookstore' instead of `library') \linebreak
\end{flushleft}
\end{enumerate}

\section{Classification of Techniques}

\begin{figure}[htbp]
	\centering
		\includegraphics[width=1.00\textwidth]{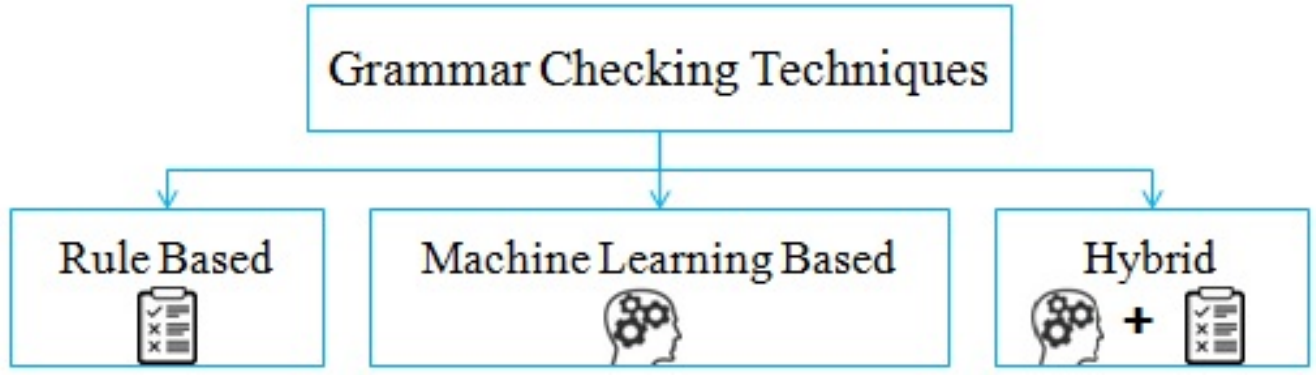}
	\caption{Classification of Grammar Checking Techniques}
	\label{fig:Gchecktechniques}
\end{figure}

This section will address our research questions RQ3 and RQ4. See figure~\ref{fig:Gchecktechniques}. There are three main techniques of grammar checking:

\subsection{Rule based technique:}
The classical approach of grammar checking is to manually design grammar rules as shown in \cite{naber2003rule}. These High quality rules are designed by linguistic experts. An English text tagged with parts-of-speech (\emph{henceforth POS}) is checked against the defined set of rules and a matching rule is applied to correct any error. The technique appears to be simple as it is easy to add, edit or remove a rule; however, writing rules needs extensive knowledge of the underlying language's grammar. Rule based systems can provide detailed explanation of flagged errors thus making the system extremely helpful for the purpose of computer aided language learning. But manual maintenance of hundreds of grammar rules is quite tedious.

\subsection{Machine Learning based technique:}
Machine learning is currently the most popular technique of grammar checking. A method that uses supervised learning provides best results \cite{sidorov2013rule}. These methods use an annotated corpus which in turn is used to perform statistical analysis on the text to automatically detect and correct grammar errors. Unlike rule based systems, it is difficult to explain the errors resulted by these systems. Machine learning based systems does not require extensive knowledge of the grammar since it is completely dependent on the underlying corpus. Non-availability of a large annotated corpus hinders the application of such techniques for grammar checking purpose. Also, the results greatly depend on how clean the corpus is.

\subsection{Hybrid technique:}
A combination of machine learning and rule based techniques can be utilized to improve the performance of the system. Since, some errors are better solved by rule based technique (e.g., use of `a' or `an') and some are better solved by machine learning (e.g., determiner errors). So, each part of the hybrid technique should be implemented according to its `competence' \cite{sidorov2013rule}. As experimented in \cite{chodorow2007detection}, the corpus of text can be used to train the system for identifying correct pattern of sentences and the results can be filtered by applying some hand-crafted rules. Hybrid technique is helpful in addressing a wide range of complex errors. Also, the tedious job of writing so many rules can be reduced to a greater extent.

\section{Literature Review}

In this section, we present our study of various approaches that we have selected as our primary studies. For each primary study, we explain the approach, give a graphical representation of it, discuss the types of error that can be detected or corrected by it, discuss about the experiments and results that are presented by the authors in their respective papers and discuss the strengths and limitations of the approach. This section will address RQ5, RQ6 and RQ10. RQ7 and RQ8 will be addressed by table~\ref{tab:comparison2} and table~\ref{tab:comparison1}.

\subsection{English Grammar Checker (1997):} 
\begin{figure}[htbp]
	\centering
		\includegraphics[width=1.00\textwidth]{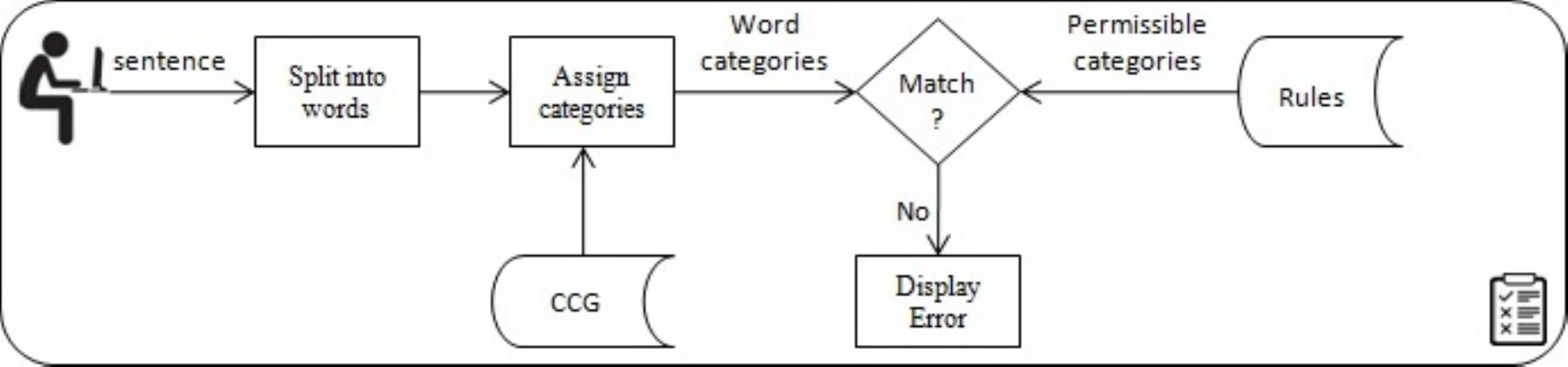}
	\caption{Schematic Diagram of English Grammar Checker \cite{park1997english}}
	\label{fig:EngGramCheck}
\end{figure}

Park et al\cite{park1997english} developed a web interface named English grammar checker which aims at the detection of grammatical errors commonly made by university students. The approach utilizes Combinatory Categorial Grammar (CCG) to derive syntax information of a sentence in a categorical lexicon. Each categorical lexicon is a collection of lexical entries. An entry is a kind of rule which defines acceptable categories of words that are local to a given word. For example, for article `a', the entry would describe that after article `a', category NP (third person singular) is expected and further category VP (compatible with NP) can be expected to form a sentence. If a sentence derivation violates such rule, associated error message is displayed. The authors have tested the approach for identifying errors made by students of University of X in their English essays. See figure~\ref{fig:EngGramCheck}\\

This is a purely syntactic approach, where grammar errors concerning the wrong syntax of a sentence can be detected. A sentence is rejected, if its derivation is not acceptable. To accept a sentence, simply add a new entry in the lexicon. The approach is able to detect spelling errors, article or determiner error, agreement errors, missing or extra elements and verb tense error. All other type of errors such as wrong word choice errors, preposition errors and run-ons could not be detected. Also, it reported some level of misdiagnoses. This interface in currently not available on the web.

\subsection{Island processing based Approach (1997):} 
\begin{figure}[htbp]
	\centering
		\includegraphics[width=1.00\textwidth]{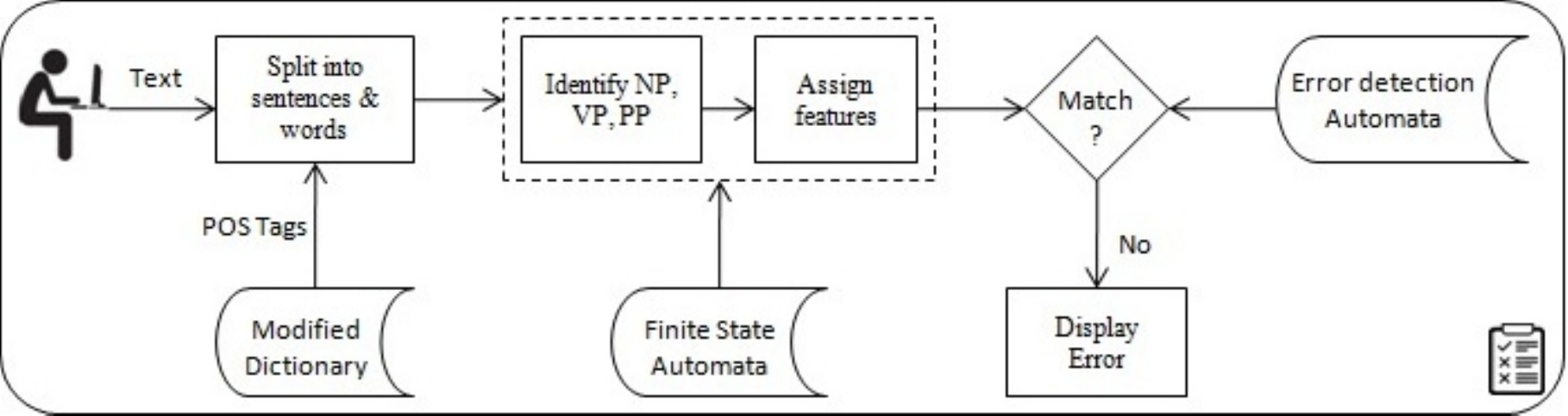}
	\caption{Schematic Diagram of Island Processing based Approach \cite{tschichold1997developing}}
	\label{fig:island}
\end{figure}

Tschumi et al \cite{tschichold1997developing} developed an English grammar checking tool for native speakers of French language using a method called island processing. The tool works in four steps. In the first step, input text is broken into sentences and words. The words are assigned a syntax category (POS tag). In the second step, a set of finite state automata is used to identify noun phrases, verb phrases and prepositional phrases as the important islands in the sentence. Depending upon the type of noun or preposition, they are assigned specific features and stored in registers. The third step calls the error detection automata which matches the word features to decide on an error and suggests a correction to it. The authors have compared their prototype with other commercial grammar checkers and have reportedly performed better. However, they did not discuss the data which they have used for comparison. See figure~\ref{fig:island}

The proposed prototype uses a scaled-down version of full dictionary which consist of words along with its syntactic category to assign POS tag instead of using a parser which saves time when parsing an ill-formed sentence. Also, island processing method lowers the processing time. To reduce overflagging, an error must be correctly identified. For this purpose the tool provides a user interaction module asking question to user and a problem word highlighter explaining problematic word usage. Though the method successfully reduces overflagging of errors, it fails at automatic correction. The tool explains an error, suggests possible corrections and often asks question to the user, which seems annoying. The tool is not available online.

\subsection{LanguageTool (2003):} 
\begin{figure}[htbp]
	\centering
		\includegraphics[width=1.00\textwidth]{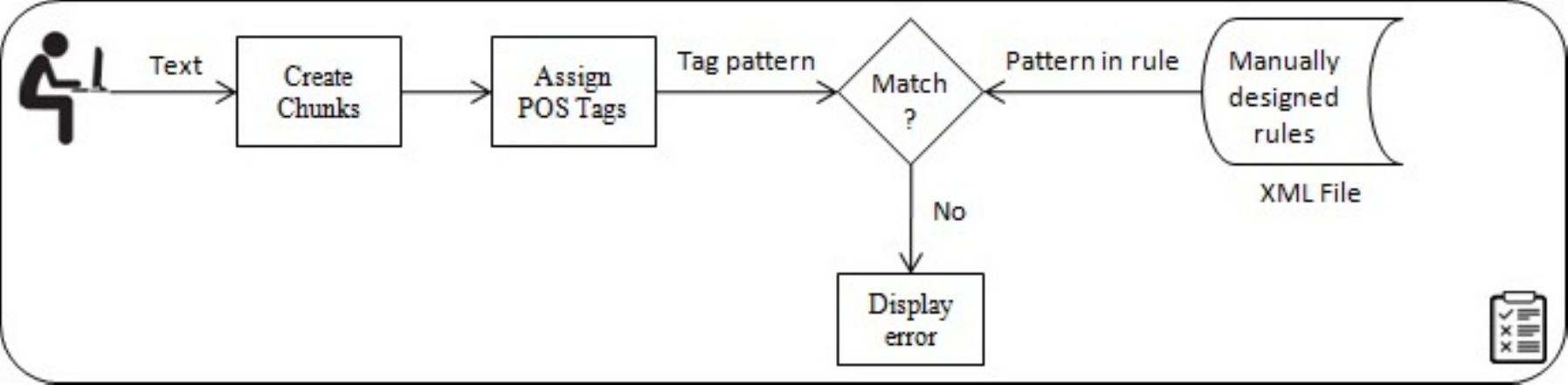}
	\caption{Schematic Diagram of LanguageTool\cite{naber2003rule}}
	\label{fig:Langtool}
\end{figure}

Naber\cite{naber2003rule} proposed LanguageTool which is an open source English grammar checker based on traditional rule based approach. The method splits the text into chunks and all the words are POS tagged. The task of spell checking is done by Snakespell python module integrated with the system. It uses a probabilistic tagger Qtag with a rule based extension for POS tagging and a rule based chunker for chunking of text into phrases. Next the manually designed XML based rules are applied to detect errors in the text. These rules define the erroneous pattern of POS tags. When applied, each rule matches the tag pattern given in the rule with the tag pattern present in text. If a match occurs, an error is detected and the system provides explanation messages and example sentences. See figure~\ref{fig:Langtool}

The author did not discuss about the data on which the tool was tested. LanguageTool is a very precise grammar checker available on the web. It can be used as a standalone web application as well as can be integrated with a text editor. It supports more than 20 languages with different number of rules. For English, it has 1614 XML rules. The rule set can be extended by simply adding the rule in the XML file. The obvious drawback of such system is the complex and time consuming task of rule development. Also, the large number of rules to cover majority of errors, results in its low recall.

\subsection{Arboretum (2004):} 
\begin{figure}[htbp]
	\centering
		\includegraphics[width=1.00\textwidth]{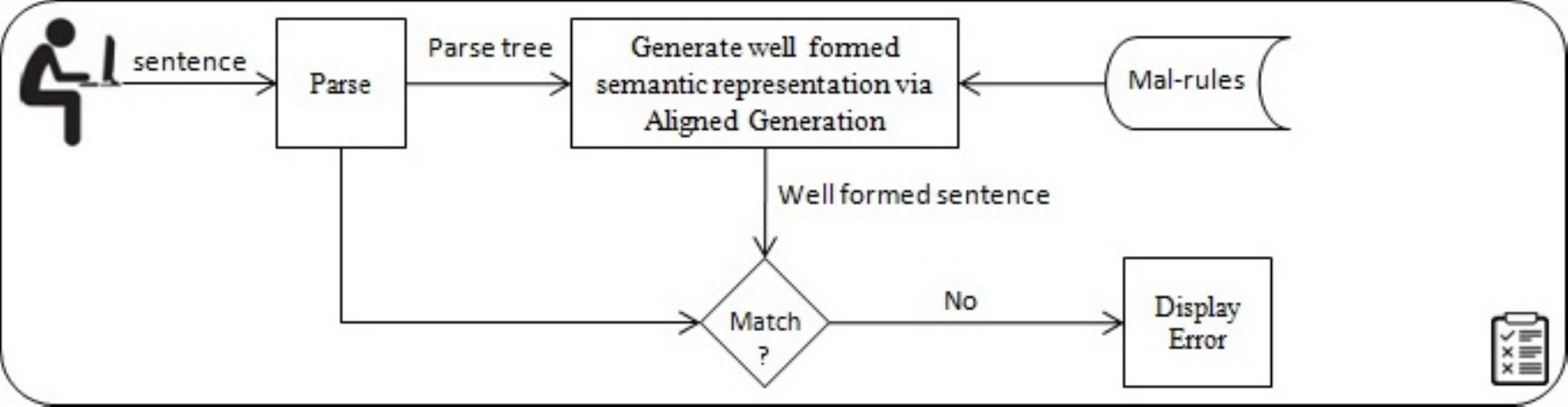}
	\caption{Schematic Diagram of Arboretum\cite{bender2004arboretum}}
	\label{fig:arboretum}
\end{figure}

Bender et al\cite{bender2004arboretum} proposed Arboretum, a tool to correct English grammar sentences based on some rules called as mal-rules. The authors have classified mal-rules in three categories namely syntactic construction mal-rules, lexical mal-rules and mal-lexical entry. The rules are then used to map correct string from incorrect one using a best-first generation method which they named as `aligned generation'. Aligned generation generates a sentence that closely matches to structure and lexical yield of some reference sentence. Priorities are assigned to the generation tasks and working up with the priorities, the first complete tree found by the generator is considered as the closest to the reference parse. See figure~\ref{fig:arboretum}

The system was tested on a sample of 221 items taken from SST corpus. The authors report that the tool is able to generate correct string in 80\% of cases of the experiment. However, the experimental dataset taken was small with an aim of finding a few types of errors. The proposed strategy failed in some cases due to its inability in identifying lexical entries and phrasal tasks. The tool is not available online, so it is not clear whether it supports automatic correction or not.

\subsection{SMT based approach (2006):} 
\begin{figure}[htbp]
	\centering
		\includegraphics[width=1.00\textwidth]{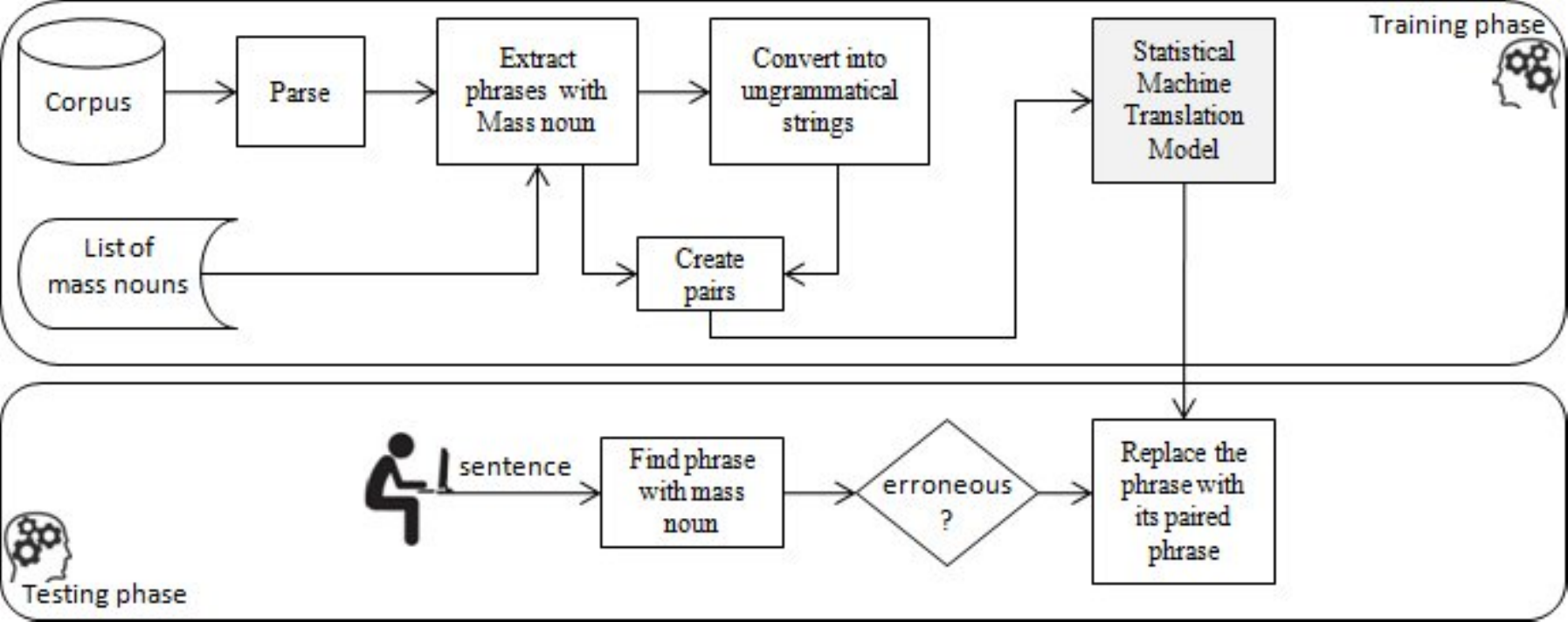}
	\caption{Schematic Diagram of SMT based Approach\cite{brockett2006correcting}}
	\label{fig:SMT}
\end{figure}

The approach proposed by Brockett et al\cite{brockett2006correcting} makes use of Statistical Machine Translation (SMT) to detect and correct grammar errors. Aiming at mass noun errors, the authors advocate translation of the whole erroneous phrase instead of individual words. A noisy channel model was used for error correction using SMT technique. The work identifies 14 nouns that frequently occurred in CLEC corpus. The sentences containing these errors are used to create training data which can map erroneous string to correct one. See figure~\ref{fig:SMT}

The system was tested on 123 example sentences taken from English websites in China. During testing, the approach was able to correct 61.81\% of mass noun errors. Errors like subject verb agreement and punctuation errors were simply ignored. The system was not able to correct an error where a word is both mass noun and count noun; for example, the word `paper' in the given two phrases- `many paper' and `five pieces of papers'.  Also the training data did not cover all other types of grammar errors made by ESL learners. This system is not available online.

\subsection{Maximum Entropy Classifier based approach (2007):} 
\begin{figure}[htbp]
	\centering
		\includegraphics[width=1.00\textwidth]{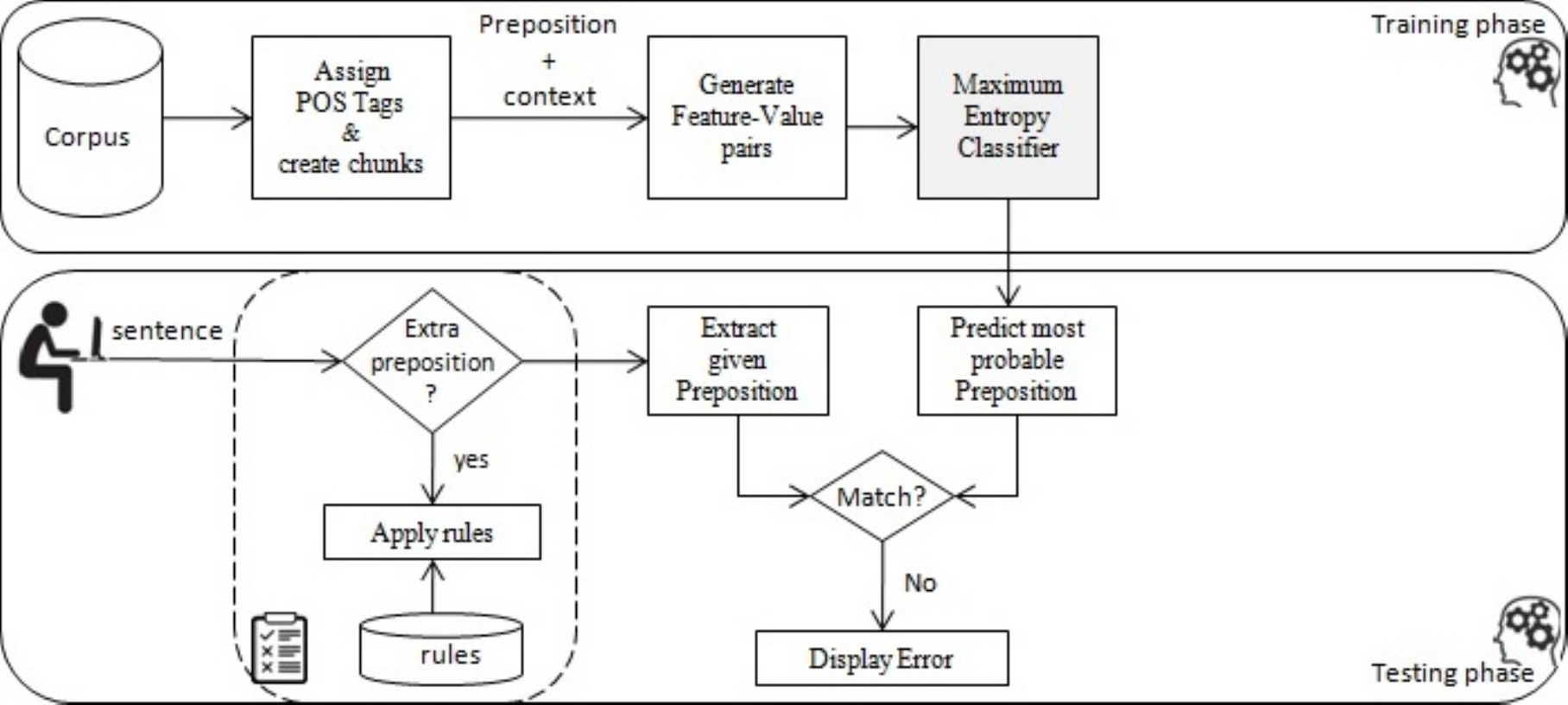}
	\caption{Schematic Diagram of Maximum Entropy Classifier based Approach\cite{chodorow2007detection}}
	\label{fig:chodorow}
\end{figure}

The approach proposed by Chodorow et al\cite{chodorow2007detection} aims at detecting prepositional errors in a corpus of ESL text. For this task a maximum entropy model is used which is trained with prepositions along with a set of associated feature-value pairs (its context). The sentences are POS-Tagged and chunked. 25 features were used to train the maximum entropy model where each feature is associated with some values. The feature-value pairs having very low frequency of occurrence were eliminated. The model is then tested on a different dataset. The model predicts the probability of each preposition in the given context and then compares it with the preposition used by the writer. The erroneous preposition is replaced with most probable preposition. Subsequently, each context is classified into one of the 34 classes of preposition. To solve the problem of detecting extra preposition, authors devised two rules- Rule1 deals with repetition of same preposition and error is detected when same POS tag is used. Rule2 deals with wrong addition of a preposition between a plural noun and a quantifier. See figure~\ref{fig:chodorow}

This approach uses a huge dataset for training and testing purpose. Training is done on 7 million prepositional contexts taken from MetaMetrics corpus and newspaper text and testing is done on 18157 prepositional contexts taken from a portion of Lexile text and 2000 contexts from ESL essays. This approach deliberately skips the contexts in the following cases- when there is a slight difference between the most probable and second most probable preposition, when adjacent words are misspelled, when there are comma errors, when the writer uses antonym of a preposition, and also in case when benefactives are used. Also, the rules for detecting extraneous preposition are insufficient to cover other types. No tool support is available for this approach.

\subsection{AIS based approach (2007):} 
\begin{figure}[htbp]
	\centering
		\includegraphics[width=1.00\textwidth]{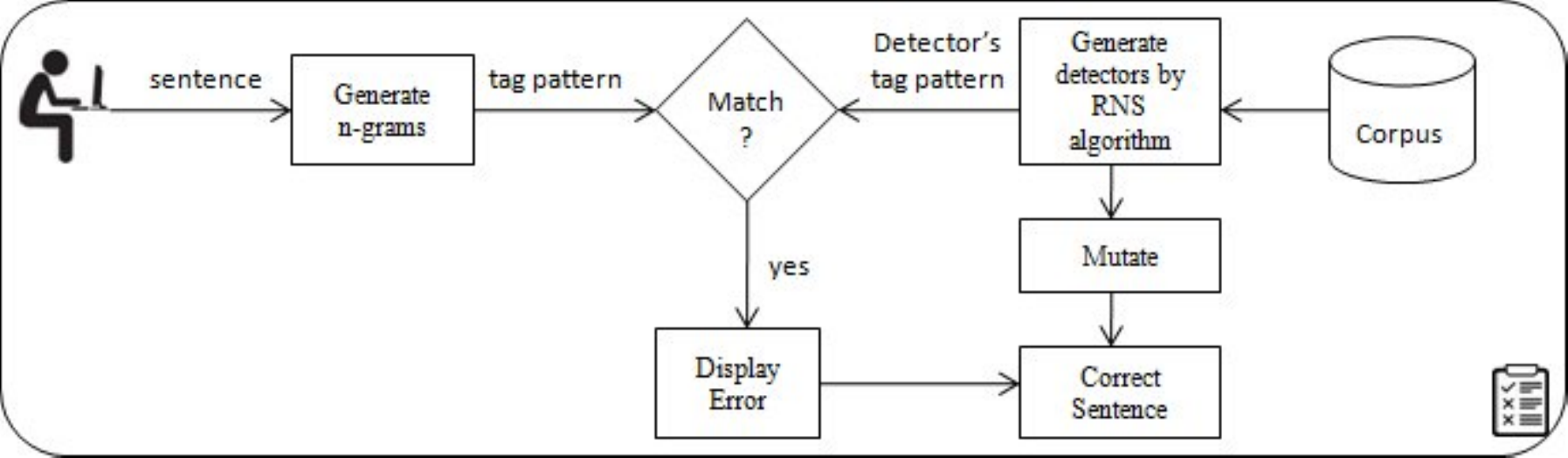}
	\caption{Schematic Diagram of AIS based Approach\cite{kumar2007artificial}}
	\label{fig:AIS}
\end{figure}

Kumar et al\cite{kumar2007artificial} proposed an approach of grammar checking, inspired from human immune system. Like the human immune system generates immune cells to detect antibodies, similarly a large corpus can be used to detect ungrammatical sentences by generating detectors. The detectors are the sentence constructs that do not appear in the corpus. A test sentence is taken to form bigrams, trigrams and tetragrams. These are tagged with extended POS tags. The sequence of tags which does not exist in the corpus is called as detector and is used to flag error. Next the detector is cloned to repair ungrammatical construct into correct one. The authors have used Real Valued Negative Selection Algorithm to generate detectors and also to fine tune the set of detectors which are capable of identifying errors better and quicker. See figure~\ref{fig:AIS}

This is a language independent approach based on Artificial Immune System (AIS) where the underlying corpus (Reuters-21578) mimics the human immune system. Any sentence construct outside the corpus is regarded as error even though it is grammatically correct. The authors have tested the system on sentences taken from a book of grammar errors. However, the size of the testing data and the results of the experiments are not discussed in their paper. This approach is able to identify 8 type of errors namely subject-verb agreement errors, wrong verb tense, adverb, adjective error, article, pronoun, wrong noun number error and missing verb error. All other types of errors are undetected. The authors argue that these shortcomings of the approach can be solved by extending the POS tag set. Still, the task of creating a corpus which is large enough to include all type of correct sentences seems practically infeasible. No tool support is available for this approach.

\subsection{LSPs based approach (2007):} 
\begin{figure}[htbp]
	\centering
		\includegraphics[width=1.00\textwidth]{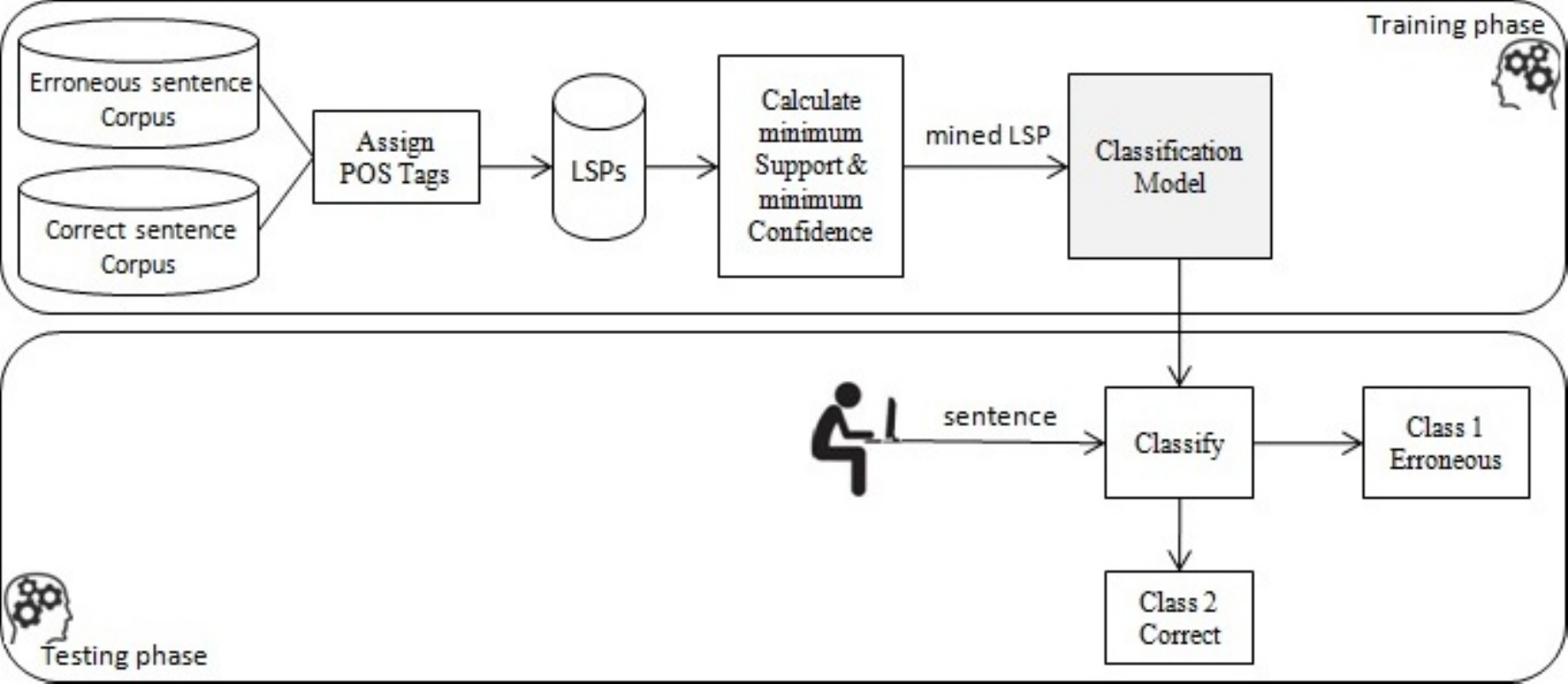}
	\caption{Schematic diagram of LSP based Approach\cite{sun2007detecting}}
	\label{fig:LSP}
\end{figure}

Sun et al\cite{sun2007detecting} proposed an approach which combines pattern discovery and machine learning to classify a sentence into two classes: correct and erroneous. To build this classification model, labeled sequential pattern (LSP) is used as an input feature. The training data is POS tagged and the frequently occurring patterns are discovered from both correct and erroneous sentences. Based on whether the pattern satisfies the given constraints for support and confidence, these patterns are labeled as erroneous or correct. Along with LSPs, other linguistic features like syntactic score, lexical collocation, function word density and perplexity are also used to detect different types of errors. See figure~\ref{fig:LSP}

The method was implemented on SVM and Bayesian classifiers using HEL, JLE and CLEC corpus. Different experiments were carried out to analyze and compare various results. The Authors found that LSP feature performs better in every case. Also they compared their method with two prototypes and it outperformed the other two in terms of precision, recall and F-score. The method can detect various grammar errors, lexical collocation errors and wrong sentence structure errors. However, the automatic correction of detected errors is not supported. Also spelling errors are simply ignored. No tool support is available for this approach.

\subsection{Auto-Editing (2010):} 
\begin{figure}[htbp]
	\centering
		\includegraphics[width=1.00\textwidth]{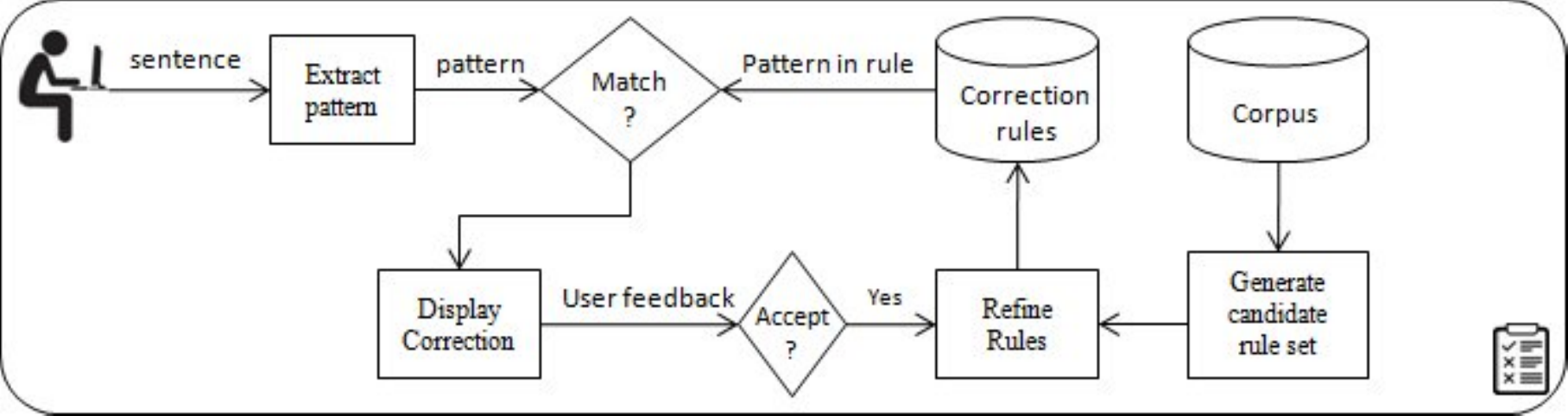}
	\caption{Schematic Diagram of Auto-Editing\cite{huang2010discovering}}
	\label{fig:autoediting}
\end{figure}

Huang et al\cite{huang2010discovering} developed an online tool for automatic grammar error correction. The approach uses a manually created corpus of paired sentences collected from a website lang-8.com. A pair consists of an erroneous sentence and its respective correct sentence. Then the corpus is used to derive sentence correction rules. A rule is represented as A→B where A is a word pattern found in erroneous sentence and B is the pattern found in its respective correct sentence. These patterns are identified by calculating Edit distance at word level. An edit distance (Lavenstein distance) is the minimum number of insert, delete or substitute operations required to transform erroneous pattern to correct pattern.  This would result in the generation of candidate rules. Among these, the rule which is able to transform erroneous sentence of the corpus to its correct form is applied while others are ignored. See figure~\ref{fig:autoediting}

This approach is an application of pattern mining on English sentences. In this approach, the rules are automatically derived from the corpus itself. The candidate rule set is refined using a condensing algorithm and ranking the rules based on user feedback (frequently used rules are top ranked). Though it achieves better precision and recall when compared with ESL assistant and Microsoft Word 2007 grammar checker, it detects mostly spelling and phrasal errors and does not cover other types of grammar errors like run-on sentences. The demo webpage of auto-editing is currently not available.

\subsection{ASO based approach (2011):} 
\begin{figure}[htbp]
	\centering
		\includegraphics[width=1.00\textwidth]{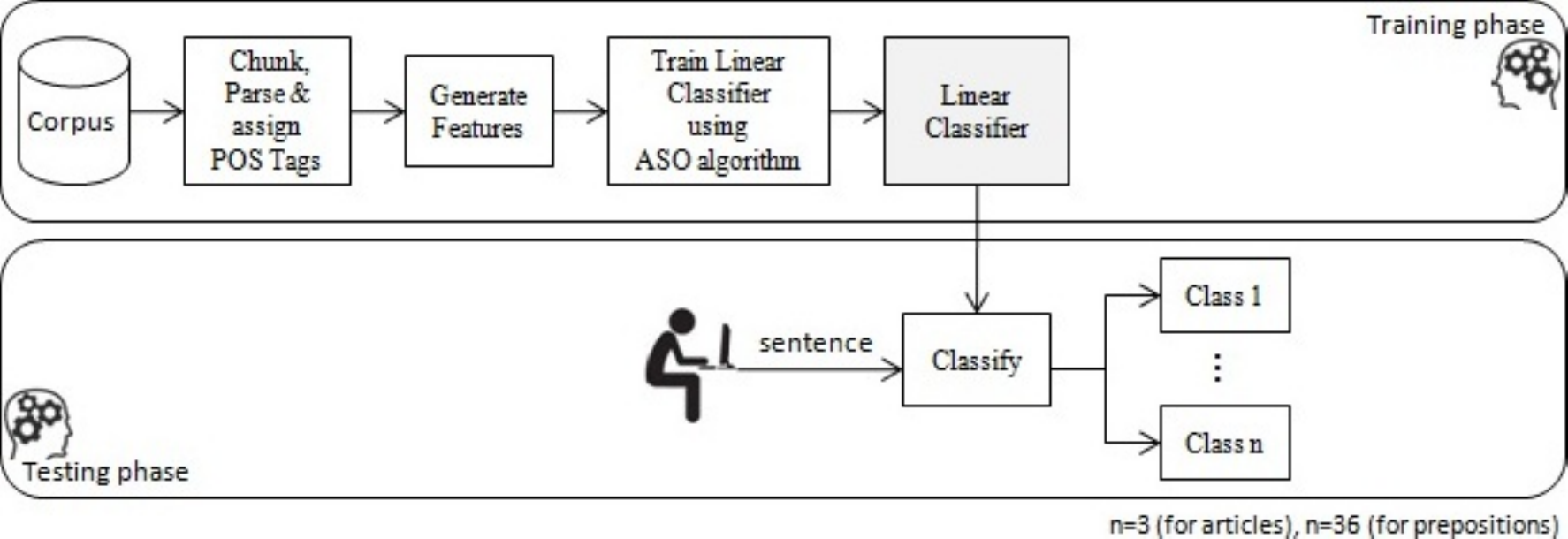}
	\caption{Schematic Diagram of ASO based Approach\cite{dahlmeier2011grammatical}}
	\label{fig:ASO}
\end{figure}

Dahlmeier et al\cite{dahlmeier2011grammatical} proposed grammar error correction using a linear classifier. They aim at correction of article and prepositional errors. The article or preposition and its context is treated as feature vectors and the corrections are treated as the classes. They used a combination of learner and non-learner text for training. When training is done on learner text, the context of article or preposition is treated as feature vector and the correct class is provided by human annotator. When training is done on non-learner text, the observed article or preposition is also added into the feature set and the correct class is the observed article or preposition. The classifier is trained using Alternating Structure Optimization (ASO) algorithm. ASO algorithm is learning the common structure of multiple related problems. This common structure can be learned by creating auxiliary problems. Auxiliary problems are helpful in predicting the wrong article or determiner in the user text. Then the classifier can be trained for these auxiliary problems to classify articles into 3 classes and prepositions into 36 classes. See figure~\ref{fig:ASO}

The training was done on NUCLE corpus and Gigaword corpus and testing was done using Wall Street Journal. The results were compared with two baseline methods. The ASO model outperforms both with an F1 measure of 19.29\% for articles and 11.15\% for prepositions. They also compared the ASO method with two commercial grammar checking tools. The performance of ASO method was far better than both the tools. There is a large scope of improvement in the performance as the precision and recall values are much less. Also, the problem of unidentified errors and false flags still persists in the system. No tool support is currently available for this approach.

\subsection{UI System (2013):} 
\begin{figure}[htbp]
	\centering
		\includegraphics[width=1.00\textwidth]{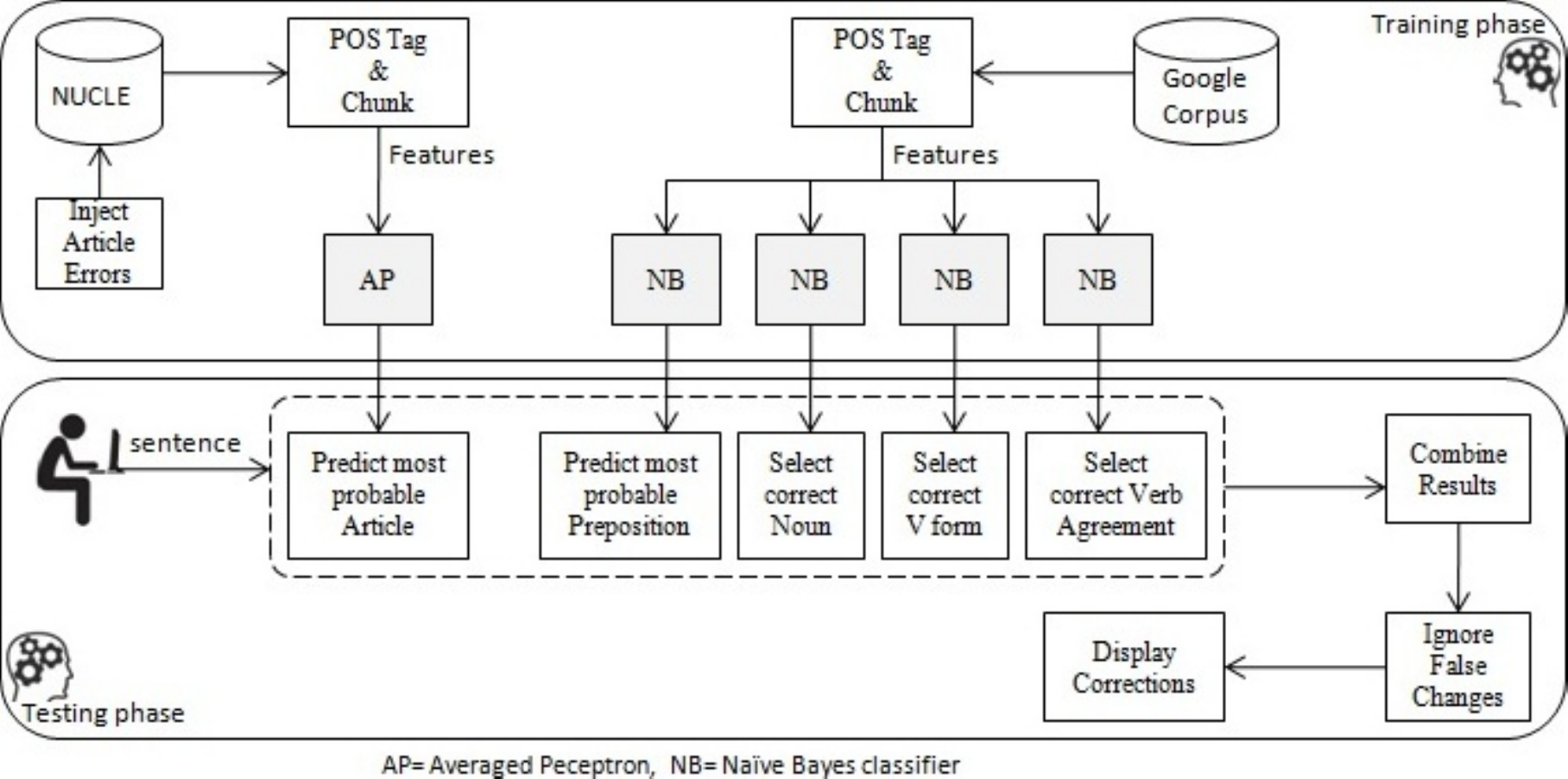}
	\caption{Schematic Diagram of UI System\cite{rozovskaya2013university}}
	\label{fig:UI}
\end{figure}

This system was developed by Rozovskaya et al\cite{rozovskaya2013university} at CoNLL-2013 shared task which aims at correction of five types of errors namely article/determiner, preposition, noun number, subject-verb agreement and verb form errors. The University of Illinois (UI) system is a combination of five machine learning classifier models, where each model is specialized to correct a specific type of error. To correct the article errors, Averaged Perceptron (AP) model is used, which is trained on NUCLE corpus using a rich set of features generated by POS tagger and chunker. Artificial article errors were introduced in the NUCLE corpus to reduce the error sparseness. To correct all other types of errors, Naïve Bayes (NB) classifier is trained with Google web 1T 5-gram corpus using word n-gram features. Each individual model predicts the most probable word from its candidate set.  The candidate set for articles and prepositions are (a,the,\(\phi\)) and 12 most frequent prepositions, respectively. For noun, verb agreement and verb form errors the candidate set includes their respective morphological variants. Finally, the results of each individual classifier are combined, filtered for false alarms and then applied to correct the sentence. See figure~\ref{fig:UI} 

This system was later extended in CoNLL-2014 shared task where it was implemented to correct more types of errors and to address correction of two or more related errors by using joint inference method \cite{rozovskaya2014illinois}. Though, the system performed best in the given task, the recall (8.81) and F1-score (14.84) are low for preposition errors. The developed system is not available online.

\subsection{Hybrid System (2014):} 
\begin{figure}[htbp]
	\centering
		\includegraphics[width=1.00\textwidth]{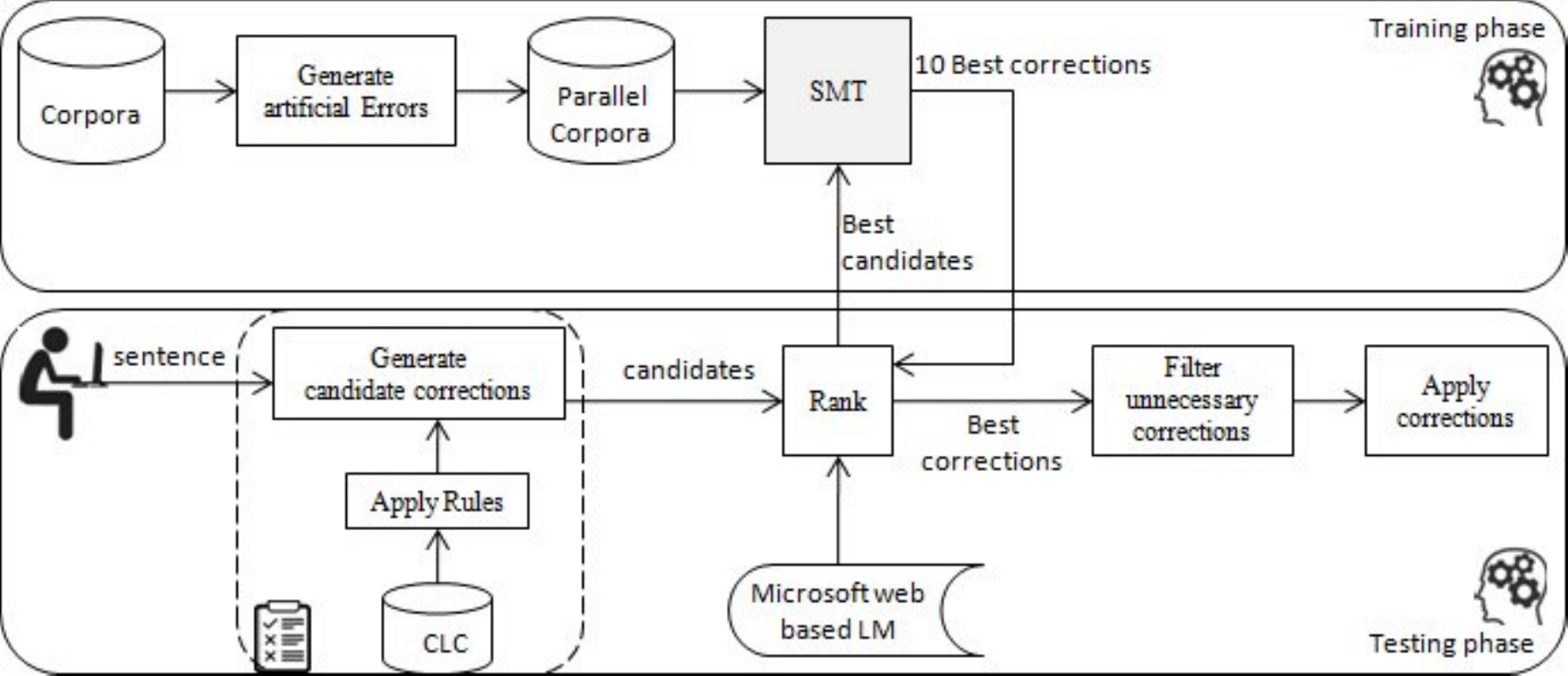}
	\caption{Schematic Diagram of Hybrid System\cite{felice2014grammatical}}
	\label{fig:CAMB}
\end{figure}

The hybrid system was developed by Felice et al\cite{felice2014grammatical} at CoNLL-2014 shared task which combines a rule based and statistical machine translation systems in a pipeline. The rule based module automatically derives rules from Cambridge Learner Corpus (CLC) that detects erroneous unigram, bigrams or trigrams and generates a list of candidate corrections. These candidates are ranked (most probable correction is top ranked) using a Language Model (LM) built from Microsoft's web n-grams. The results of the LM are pipelined into the Statistical Machine Translation system. The SMT model was trained on multiple corpora including NUCLE v3.1, 2014 shared task dataset, IELETS dataset from CLC corpus, EVP corpus and FCE corpus. The SMT model generates 10 best correction candidates which are further ranked by the language model. Next, the unnecessary corrections (corrections having error types: reordering, word acronym or run-ons) are filtered out and best correction is applied to replace the input string. See figure~\ref{fig:CAMB}

The SMT system is developed using best performing tools like Pialign for word alignment, IRSTLM to build target language model and Moses for decoding. The system is able to achieve best values for precision, recall and F-score. It is best suitable for correcting agreement errors, verb form errors, noun number, pronoun reference, punctuation, capitalization and spelling errors. However, it performs poorly on sentence fragments, run-ons, word reordering and collocations errors. The developed system is not available online.

\begin{table}[ht]
\caption{Errors detected by various Grammar Checking approaches}
\centering
\begin{tabular}{|c|c|c|c|c|c|c|c|c|c|c|c|c|c|c|}
\hline  \rotatebox{90}{Approach}
& \rotatebox{90}{Sentence structure}
& \rotatebox{90}{Fragments}
& \rotatebox{90}{Run-ons}
& \rotatebox{90}{Spelling}
& \rotatebox{90}{Syntax errors}
& \rotatebox{90}{S-V Agreements}
& \rotatebox{90}{Vform, Vtense}
& \rotatebox{90}{Noun Number}
& \rotatebox{90}{Art or Det}
& \rotatebox{90}{Preposition}
& \rotatebox{90}{Punctuation}
& \rotatebox{90}{Semantic Errors}
& \rotatebox{90}{Contextual}
& \rotatebox{90}{Word choice} \\ \hline

\cite{park1997english} & & {\cmark} & & {\cmark} &  & {\cmark} & {\cmark} &  & {\cmark} & & & & & \\ \hline

\cite{tschichold1997developing} & {\cmark} &  &  & {\cmark} & & {\cmark} & {\cmark} & {\cmark} & & & & & & {\cmark} \\ \hline

\cite{naber2003rule} & & &  & {\cmark} & {\cmark} & {\cmark} & &  & {\cmark} & & {\cmark} & {\cmark} & {\cmark} & \\ \hline

\cite{bender2004arboretum} & & &  & &  & & {\cmark} & {\cmark} & {\cmark} & & & & & {\cmark} \\ \hline

\cite{brockett2006correcting} & & & & &  & & & {\cmark} & & & & & & \\ \hline

\cite{chodorow2007detection} & & & & & & & & & & {\cmark} & & & & \\ \hline

\cite{kumar2007artificial} & & & & & & {\cmark} & {\cmark} & {\cmark} & {\cmark} & & & & & \\ \hline

\cite{sun2007detecting} & {\cmark} & {\cmark} & & {\cmark} & & {\cmark} & {\cmark} &  & {\cmark} & & & & & \\ \hline

\cite{huang2010discovering} & & & & {\cmark} & & {\cmark} & {\cmark} & & {\cmark} & {\cmark} & {\cmark} & & & \\ \hline

\cite{dahlmeier2011grammatical} & & & & & & & & & {\cmark} & {\cmark}&  &  &  & \\ \hline

\cite{rozovskaya2013university} & & & & & & {\cmark} & {\cmark} &  {\cmark} & {\cmark} & {\cmark}& & & & \\ \hline

\cite{felice2014grammatical} & & & & {\cmark} &  & {\cmark} & {\cmark} & {\cmark} & {\cmark} & {\cmark} & {\cmark} &  &  &  \\ \hline

\end{tabular}
\label{tab:comparison2}%
\end{table}%

\begin{footnotesize}
\begin{longtable}{p{4.2em}p{4.7em}p{8.5em}p{8.0em}p{8.5em}p{8.5em}p{8.5em}} 
\caption{Summary of Various Grammar Checking Approaches}
\label{tab:comparison1} \\
\toprule \toprule
\textbf{Approach} & \textbf{Technique used} & \textbf{Target error types} & \textbf{Linguistic data used} & \textbf{Results} & \textbf{Strengths} & \textbf{Limitations} \\
\midrule
\endfirsthead
\toprule
& \textbf{Technique used} & \textbf{Target error types} & \textbf{Linguistic data used} & \textbf{Results} & \textbf{Strengths} & \textbf{Limitations} \\
\endhead

\center\cite{park1997english} & \raisebox{-\totalheight}{\includegraphics{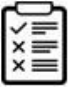}} & wrong capitalization, agreement errors, Vform error, missing fragments. & Essays written by university students. & Not specified & Simple, Customizable to identify frequent errors. & Handmade rules, No automatic correction. \\ \hline

\center\cite{tschichold1997developing} & \raisebox{-\totalheight}{\includegraphics{figures/ruleicon.pdf}} & Spelling error, S-V-A, Vtense, word choice errors, sentence error, noun error. & Self created corpus of 27000 words of text by French native speakers. & Not specified & Reduced processing time for tagging, Reduced error overflagging & Overflagging is still high, 43.5\% \\ \hline
		
\center\cite{naber2003rule} & \raisebox{-\totalheight}{\includegraphics{figures/ruleicon.pdf}} & Punctuation, syntax, semantic, style error. & Mailing list error corpus of 224 sentences. & Not specified & Simple, Large set of rules, Easy rule addition & No automatic correction, difficult to manage large number of handmade rules.\\ \hline
				
\center\cite{bender2004arboretum} & \raisebox{-\totalheight}{\includegraphics{figures/ruleicon.pdf}} & Determiner errors, noun number error, verb tense error, word choice, vform error. & SST corpus of 221 sentences & Success rate = 80\% & Better error correction due to best first method used. & Overflagging of errors, poor performance for S-V agreement errors, missing auxiliary, complement and vform errors. \\ \hline

\center\cite{brockett2006correcting} & \raisebox{-\totalheight}{\includegraphics{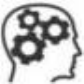}} & Mass noun errors & Reuters newswire articles, CLEC corpus. English sentences from Chinese websites. & success rate = 61.81\% & Automatic correction & Unable to detect when mass noun is also a count noun. \\ \hline
	
\center\cite{chodorow2007detection} & \raisebox{-\totalheight}{\includegraphics{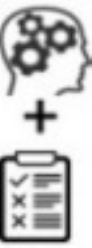}} & Preposition errors & MetaMetrics corpus of 1100 \& 1200 Lexile text, newspaper text, Chinese, Japanese \& Russian's ESL essays. & precision = 0.8, recall = 0.304 & Provides better results due to hybrid approach. & Insufficient number of rules, Low recall, Many errors were deliberately skipped. \\ \hline

\center\cite{kumar2007artificial} & \raisebox{-\totalheight}{\includegraphics{figures/ruleicon.pdf}} & SVA, article, noun number, verb error, wrong adjective, adverb or pronoun. & Reuters-21578 corpus, sentences from book- Avoid Errors by A.K. Misra. & Not specified & Language independent method, Quicker response by frequent detectors. & Any pattern outside corpus is flagged as error even if it is correct. \\ \hline
		
\center\cite{sun2007detecting} & \raisebox{-\totalheight}{\includegraphics{figures/mlicon2.pdf}} & Syntax errors, word choice error, sentence structure error. & Hiroshima English Learners corpus, Japanese Learners of English corpus \& Chinese Learner Error corpus. & Accuracy = 81.3 , precision = 83.09 , recall = 81.24 , Fscore = 81.25 & Good feature set, provides better error detection. & Does not detect spelling error. \\ \hline

\center\cite{huang2010discovering} & \raisebox{-\totalheight}{\includegraphics{figures/ruleicon.pdf}} & Spelling, phrases, SVA, punctuation, article, preposition, verb tense, gerund misuse \& other POS errors. & Self created corpus of incorrect and correct sentences collected from lang-8.com. &  Success rate = 67.2\% Precision = 40.16\% , Recall = 20.28\% & Automatic rule generation. & Most of the detected errors are spelling or phrasal errors. \\ \hline

\center\cite{dahlmeier2011grammatical} & \raisebox{-\totalheight}{\includegraphics{figures/mlicon2.pdf}} & Article and preposition errors & NUCLE corpus, Gigaword Corpus, section 23 of Wall Street Journal.& F1 = 19.29\% (articles), F1 = 11.15\% (Prepositions) & Supports automatic correction, Performance is better than commercial tools. & False flags. Unidentified errors, Low recall \& precision. \\ \hline

\center\cite{rozovskaya2013university} & \raisebox{-\totalheight}{\includegraphics{figures/mlicon2.pdf}} & Noun number, agreement error, Vform, ArtOrDet, Prepopsition errors. & NUCLE, Google web 1T 5-gram corpus. & Precision = 62.19 , Recall = 31.87 F1 , score = 42.14 & Supports automatic correction, Best performance on the target error types. & Low recall for prepositions, Inconsistent predictions due to globally interacting errors. \\ \hline

\center\cite{felice2014grammatical} & \raisebox{-\totalheight}{\includegraphics{figures/hyb.pdf}}& Total 28 types of errors \cite{felice2014grammatical} & NUCLE, CLC, FCE, EVP \& CoNLL-2014 task dataset. & Precision = 46.70 , Recall = 34.30 , F0.5 score = 43.55 & Supports automatic correction, Best performance on punctuation, spelling, capitalization, noun number, Vform, agreement errors. & Cannot handle fragments, run-ons, acronyms, idioms, word reordering \& collocation errors. \\ \hline \hline

\end{longtable}
\end{footnotesize}

\section{Conclusions and Future Research}

Grammar checking is a major part of Natural Language Processing (NLP) whose applications ranges from proofreading to language learning. Much work has been done for the development of grammar checking tools in the past decade. However, fewer efforts are made for surveying the existing literature. Thus, we present a comprehensive study of English grammar checking techniques highlighting the capabilities and challenges associated with them. Also, we systematically selected, examined and reviewed 12 approaches of Grammar checking. The 12 approaches can be classified into three categories namely (1) Rule based technique, (2) Machine learning based technique, and (3) Hybrid technique. Each technique has its own advantages and limitations. Rule based techniques are best suited for language learning but rule designing is a laborious task. Machine learning alleviates this labor but it is dependent on the size and type of the corpus used. Hybrid technique combines the best of both techniques but each part of the hybrid technique should be implemented according to its suitability.\\

In this paper, we have also presented an error classification scheme which identifies five types of errors namely sentence structure errors, punctuation errors, spelling errors, syntax errors, and semantic errors. These errors are further subcategorized. This classification scheme would help the researchers and developers in following ways: (1) identifying the most frequent errors would tell what type of errors must be targeted for correction, (2) identifying the level of the error would tell what length of text should be examined to detect any error, (3) identifying the cause of invalid text would help in finding a solution to write a valid text. This simplifies the task of grammar checking.\\

Based on our detailed review of various approaches, our observations are as follows: (1) No existing approach is completely able to detect all types of errors efficiently, (2) Most of the tools are not available for research or public use, (3) All approaches use different experimental data, thus it is hard to compare the results. (4) Most of the approaches have addressed syntax errors and its subtypes while very few efforts have been done to detect errors at sentence level and at semantic level. (5) Detection and correction of run-on sentences is yet another untouched research area, (6) No tools are suitable for real time applications like proofreading of technical papers, language tutoring, writing assistance etc. (7) Our research question RQ9 is still unanswered since we could not check the results of individual error types against gold standards, (8) Although, performance of tools has been improved gradually with time; there is much scope for improvements.\\

Based on our observations, we suggest the following emerging research directions:\\

\textbf{Classification of errors:} As noted in our study, there is a lack (or even absence) of a general classification scheme to identify types of errors. We motivate further research to suggest more classification schema. This will simplify the task of grammar checking by identifying how to handle a particular type of error.\\

\textbf{Evaluation on a standard test data:} Since all the previous approaches have been evaluated on a different test set, it is difficult to compare their performances. A standard test set of erroneous sentences with their well defined correct forms will benefit in reporting which systems are more efficient and robust.\\

\textbf{Analysis based on types of errors:} Based on our review, all the previous approaches deal with a different set of error types to be corrected. An annotated corpus which labels erroneous sentences into one of the five types and their subtypes, and then study of the performance of various approaches on each of these types will be helpful. This will help in identifying the best method for handling a specific error type.\\

\textbf{Coverage of different types of errors:} From the data in table 2, we observed that current approaches are limited in handling all types of errors, specifically sentence structure errors and semantic errors. Future work may focus on these areas.

\bibliographystyle{ACM-Reference-Format}
\bibliography{A-Systematic-Review-of-Automated-Grammar-Checking-in-English-Language}

\end{document}